%% file: main.tex
\documentclass[10pt,twocolumn,letterpaper]{article}

\usepackage{iccv}      %
\usepackage{graphics,mathptmx,times,amssymb}
\usepackage{amsmath,amssymb,epsfig,xcolor,graphicx,url,xspace,booktabs}
\usepackage{url,multirow,bm,bbm,breqn,rotating,color,colortbl,subcaption,pifont,colortbl,enumitem}

\definecolor{iccvblue}{rgb}{0.21,0.49,0.74}
\usepackage[pagebackref,breaklinks,colorlinks,allcolors=iccvblue]{hyperref}

\def\paperID{*****} %
\def\confName{ICCV}
\def\confYear{2025}

\definecolor{Gray}{gray}{0.86}

\makeatletter
\g@addto@macro\@maketitle{
  \captionsetup{type=figure}\setcounter{figure}{0}
  
  \centering
  \vspace{-2ex}
    \includegraphics[width=1.0\textwidth]{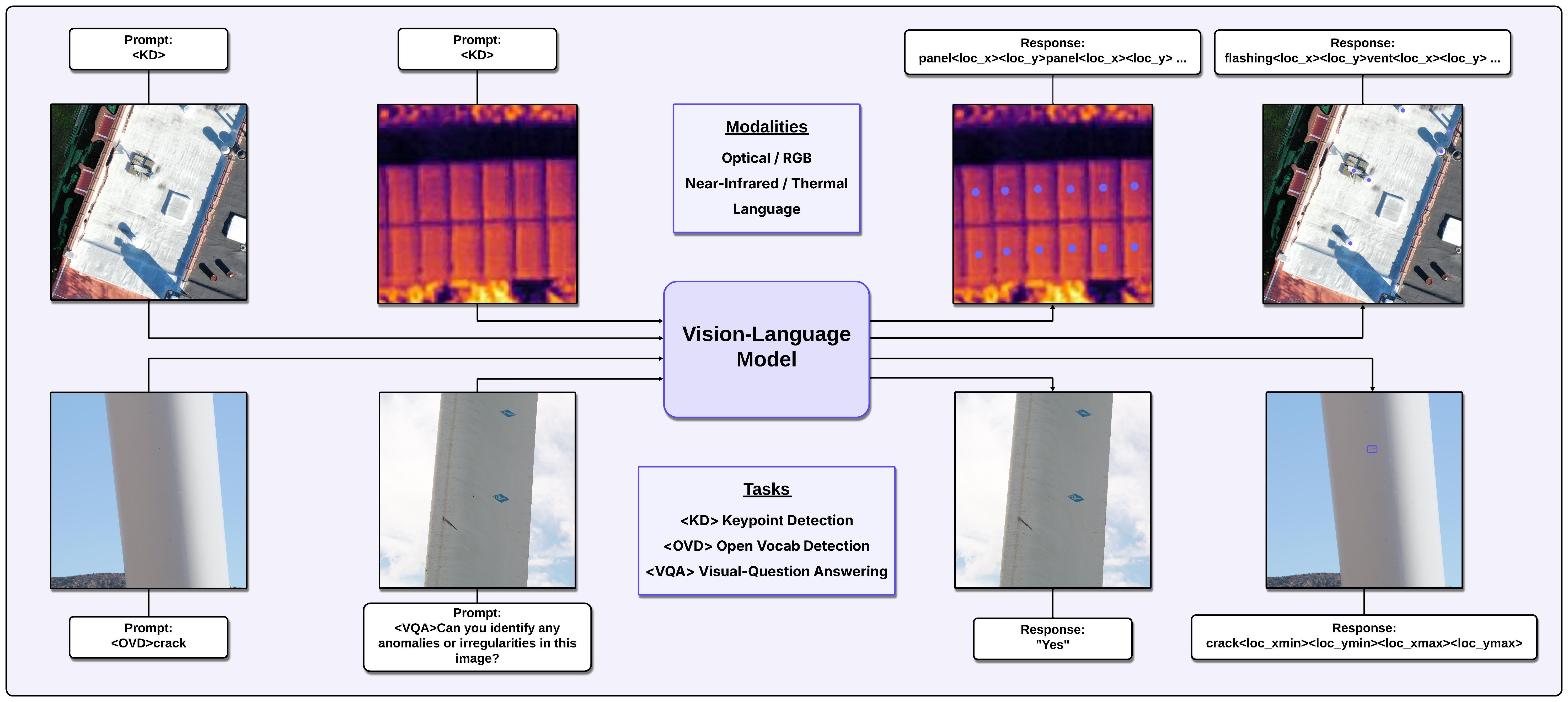}
	\caption{\textbf{Our InspectVLM multimodal multitask drone anomaly inspection architecture.} Industrial asset inspections involve multiple modalities (RGB \& Thermal imagery) and multiple tasks such as image classification, object detection, and keypoint detection. By treating visual tasks as language, Vision-Language Models (VLMs) enable the unification of independent task-specific models into a single architecture.}
    \label{fig:pipeline}
    \vspace{4ex}
}
\makeatother

\title{InspectVLM: Unified in Theory, Unreliable in Practice}

\author{Conor Wallace\thanks{Corresponding authors: \tt\footnotesize conor.wallace@zeitview.com}\\Zeitview
\and
Isaac Corley\\Zeitview
\and
Jonathan Lwowski\\Zeitview
}

\begin{document}
\maketitle
\input{ICCV2025-Author-Kit-Feb/sec/abstract}    
\input{ICCV2025-Author-Kit-Feb/sec/intro}
\input{ICCV2025-Author-Kit-Feb/sec/related_work}
\input{ICCV2025-Author-Kit-Feb/sec/methods}
\input{ICCV2025-Author-Kit-Feb/sec/data}
\input{ICCV2025-Author-Kit-Feb/sec/experiments}
\input{ICCV2025-Author-Kit-Feb/sec/results}
\input{ICCV2025-Author-Kit-Feb/sec/conclusion}

{
    \small
    \bibliographystyle{ieeenat_fullname}
    \bibliography{main}
}

\end{document}

%% file: ICCV2025-Author-Kit-Feb/sec/abstract.tex
\begin{abstract}
Unified vision-language models (VLMs) promise to streamline computer vision pipelines by reframing multiple visual tasks—such as classification, detection, and keypoint localization—within a single language-driven interface. This architecture is particularly appealing in industrial inspection, where managing disjoint task-specific models introduces complexity, inefficiency, and maintenance overhead. In this paper, we critically evaluate the viability of this unified paradigm using InspectVLM, a Florence-2–based VLM trained on InspectMM, our new large-scale multimodal, multitask inspection dataset. While InspectVLM performs competitively on image-level classification and structured keypoint tasks, we find that it fails to match traditional ResNet-based models in core inspection metrics. Notably, the model exhibits brittle behavior under low prompt variability, produces degenerate outputs for fine-grained object detection, and frequently defaults to memorized language responses regardless of visual input. Our findings suggest that while language-driven unification offers conceptual elegance, current VLMs lack the visual grounding and robustness necessary for deployment in precision-critical industrial inspections.
\end{abstract}

%% file: ICCV2025-Author-Kit-Feb/sec/intro.tex
\section{INTRODUCTION}
\label{sec:intro}

Large-scale industrial asset inspection—across wind turbines, solar farms, and building rooftops—relies on a range of computer vision tasks, including anomaly detection, object localization, and inventory counting. Traditionally, these tasks are addressed by training and deploying separate models for classification, detection, and keypoint localization, each tailored to a specific data modality and use case. While effective, this approach introduces substantial operational complexity and duplication: each task requires model tuning, deployment infrastructure, and long-term maintenance.

Recent advances in vision-language models (VLMs) offer an alternative. By casting vision tasks into a language interface—e.g., formulating detection as open vocabulary detection, or binary classification as visual-question-answering—VLMs promise to unify these disparate models into a single architecture. This unified approach could significantly simplify inspection systems: one model, one interface, multiple tasks.

However, this promise is largely untested in real-world, high-stakes domains like industrial inspection. Existing VLM evaluations focus on curated web benchmarks or synthetic data, leaving open the question: Can unified VLMs actually replace task-specific vision models in practical inspection pipelines?

In this paper, we present a case study addressing this question. We introduce InspectMM, a large-scale multimodal multitask dataset spanning over 290,000 drone-acquired images with expert-labeled annotations for classification, object detection, and keypoint detection across wind, solar, and property domains. Using this dataset, we train InspectVLM, a Florence-2–based model fine-tuned across all three tasks simultaneously. We then compare its performance to traditional models: ResNet-50 classifiers, Faster R-CNN detectors, and Keypoint R-CNN localizers.

Our findings highlight the trade-offs of unified VLMs in practice:
\begin{itemize}
\item InspectVLM performs competitively on image classification and structured keypoint tasks, particularly in solar panel arrays.
\smallskip

\item However, it fails significantly on fine-grained object detection, frequently producing degenerate bounding boxes.
\smallskip

\item The model exhibits brittle language behavior, overfitting to fixed prompt templates and ignoring visual input under low variability.
\smallskip

\end{itemize}

These results suggest that while VLMs are architecturally elegant and appealing in theory, current models do not meet the accuracy, reliability, or robustness required for industrial-grade inspection. We conclude that VLMs offer a valuable unification strategy—but only when paired with careful prompt design, adequate visual grounding, and fallback mechanisms for safety-critical applications.

In this work, we explore the effectiveness of VLMs for large-scale asset inspection across a range of tasks and compare their performance to that of traditional task-specific computer vision models. We summarize our contributions below:

\vspace{1ex}\hspace{-3.5ex}\textbf{Multimodal Multitask Industrial Inspection Dataset}\hspace{1mm} We develop a novel multimodal multitask dataset, \textbf{InspectMM}, for large-scale asset inspection, encompassing diverse sub-tasks across multiple image domains and asset types.

\vspace{1ex}\hspace{-3.5ex}\textbf{A Unified VLM for Multimodal Multitask Inspection}\hspace{1mm} We investigate the applicability of VLMs for performing industrial inspection across different asset types and image modalities, resulting in the development of our \textbf{InspectVLM} architecture.

\vspace{1ex}\hspace{-3.5ex}\textbf{A case study in unified model performance}\hspace{1mm} We conduct an empirical evaluation of our InspectVLM unified model against traditional computer vision models tailored to individual sub-tasks.

\vspace{1ex}\hspace{-3.5ex}\textbf{A detailed analysis of VLM failure modes}\hspace{1mm} We identify and quantify key limitations of current VLMs in inspection settings, including \textbf{(1)} overfitting to low prompt variability, \textbf{(2)} defaulting to degenerate bounding boxes, and \textbf{(3)} reliance on spatial pattern priors over visual features. These findings highlight the challenges of deploying unified VLMs in high-precision industrial tasks.

%% file: ICCV2025-Author-Kit-Feb/sec/related_work.tex
\subsection{Related Work}
\label{sec:related_work}
\textbf{VLMs}\hspace{1mm} The combination of natural language processing and computer vision architectures into VLMs has progressed rapidly in recent years, showing strong multitasking and generalization abilities. Architectures such as LLaVA~\cite{llava}, MiniGPT-4~\cite{minigpt4}, InstructBLIP~\cite{instructblip}, GroundingDINO~\cite{liu2023grounding} have shown improved performance for multimodal tasks such as image captioning, visual question answering (VQA), and open-vocabulary detection. Furthermore, VLMs have also been used for visual-grounding tasks wherein the model is capable of understanding features from a referred image location. In a broader scope, general-purpose VLMs such as Florence-2~\cite{xiao2024florence} and PaliGemma~\cite{beyer2024paligemma} are pretrained simultaneously on a combination of multimodal tasks. These advancements indicate that VLMs are a viable alternative to traditional task-specific vision models.

\input{ICCV2025-Author-Kit-Feb/figures/dataset}

\vspace{1ex}\hspace{-3.5ex}\textbf{Asset Inspection}\hspace{1mm} Traditional deep learning methods have been used to great effect in multiple visual asset inspection tasks, including wind turbine inspection~\cite{agrawal2024barely}, building rooftop measurements~\cite{corley2024zrg}, and solar farm inventory management~\cite{wallace2023solar}. VLMs have also been employed in inspection settings, although they have been limited to primarily single tasks or usage of simulated datasets. For example, AnomalyGPT~\cite{Gu_Zhu_Zhu_Chen_Tang_Wang_2024anomalyGPT} was trained on industry-specific data with simulated visual anomalies to generate descriptions of present anomalies and generate an approximate location via unsupervised learning using the feature maps generated by the vision encoder. Automotive-LLaVA~\cite{kumar2024diagnostics} was proposed for answering questions about automotive part images. Similarly, Power-LLaVA~\cite{Wang_Li_Luo_Zhu_Yang_Rong_Wang_2024powerllava} is a VQA model for power line inspections. Furthermore, while VLMs have been applied to industry-specific tasks, they have not been properly evaluated against traditional task-specific vision models.

%% file: ICCV2025-Author-Kit-Feb/figures/dataset.tex
\begin{figure*}[t]
    \centering
    \includegraphics[width=0.95\textwidth]{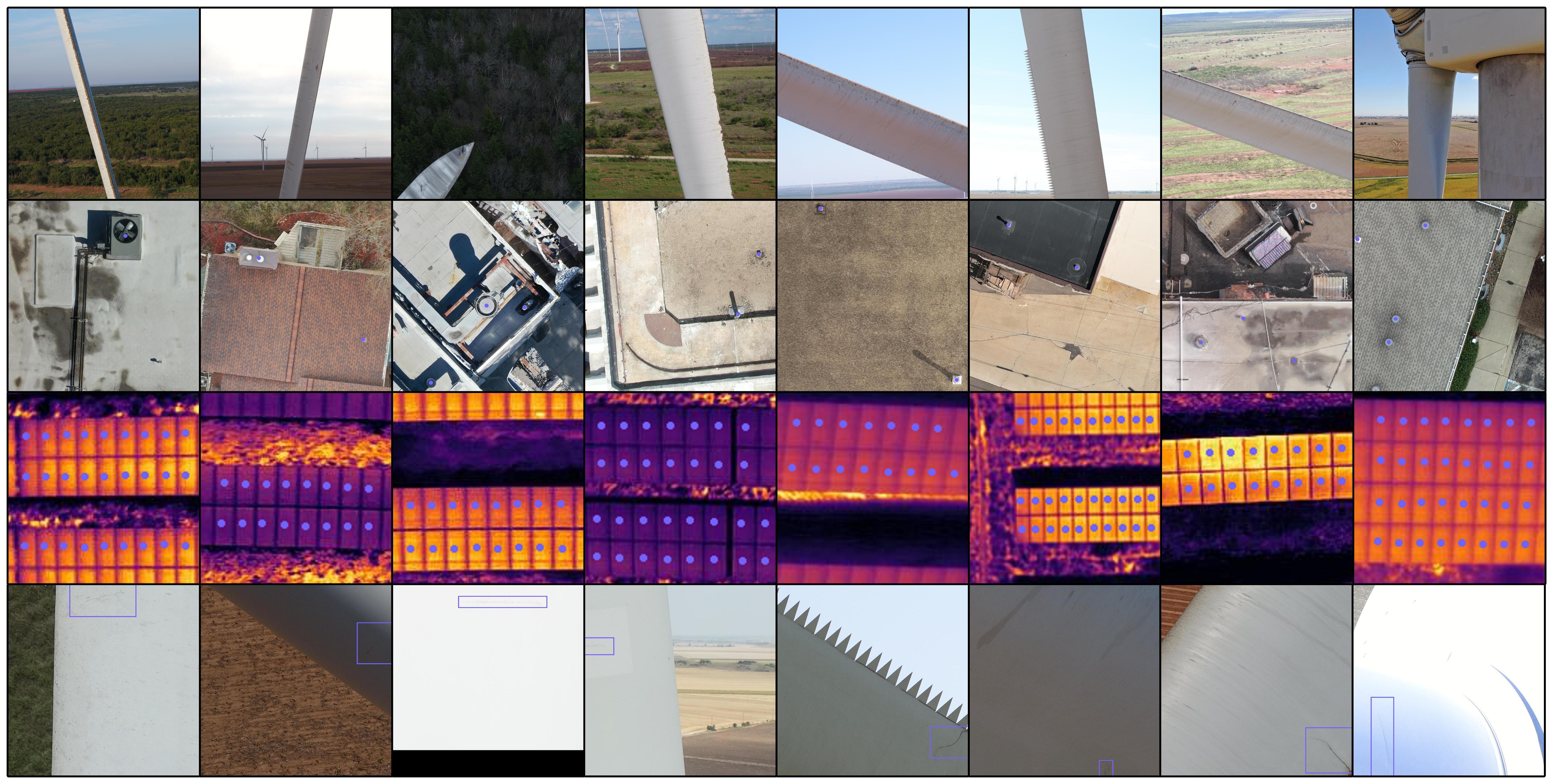}
    \caption{\textbf{Samples from our InspectMM dataset for multimodal multitask inspections.} Our dataset task types consist of keypoint detection, visual-question answering, and object detection. From top to bottom: wind turbine anomaly flagging, properties rooftop inventory counting, solar panel inventory counting, and wind turbine crack detection. \textit{Best visualized while zoomed in.}}
    \label{fig:dataset}
\end{figure*}

%% file: ICCV2025-Author-Kit-Feb/sec/methods.tex
\section{METHODS}
\label{sec:methods}

Our model follows the Florence-2 architecture~\cite{xiao2024florence} consisting of a DaViT~\cite{ding2022davit} image encoder and a unified multimodal transformer for processing both image and text tokens. Florence-2 is trained on multiple tasks including VQA, image captioning, and open vocabulary object detection. While other VLMs commonly train adapter modules to align frozen pretrained text and vision encoders' representations, Florence-2 trains both the vision multimodal encoders from scratch. This allows the model to transfer well to vision-specific tasks. Furthermore, Florence-2 employs an additional specialized tokenizer with 1,000 spatial task tokens allowing the model to encode and decode image locations. These tokens are used to represent image coordinates by normalizing pixel coordinates by image width and height and scaling by 1,000.

\subsection{Region Representation}
Similarly to the approaches described in Florence-2~\cite{xiao2024florence} and Molmo~\cite{deitke2024molmopixmoopenweights}, we represent image regions as language, employing the following spatial encoding formats:

\vspace{1ex}\hspace{-3.5ex}\textbf{Point Representation}\hspace{1mm} Points in the image are expressed as ($x_c, y_c$), where $x_c$ and $y_c$ represent the centroid coordinates of the object of interest.

\vspace{1ex}\hspace{-3.5ex}\textbf{Box Representation}\hspace{1mm} Bounding boxes are expressed as tuples of length 4: ($x_1, y_1, x_2, y_2$) corresponding to the top-left and bottom-right corners of the box.

\subsection{Task Formulation}
To create a unified multitask inspection dataset, we reformulate conventional vision tasks as language, employing and extending task-specific prompts as outlined in~\cite{xiao2024florence}. Examples of each task prompt and response can be found in Figure~\ref{fig:pipeline}:

\vspace{1ex}\hspace{-3.5ex}\textbf{Binary Classification (VQA)}\hspace{1mm} Binary classification tasks are restructured as visual-question answering (VQA) problems. Using the prompt \texttt{<VQA>}, the model is provided with a task-specific question, and the response is a binary \texttt{yes} or \texttt{no}.

\vspace{1ex}\hspace{-3.5ex}\textbf{Keypoint Detection (Pointing)}\hspace{1mm} Inspired by the pointing approach used by Molmo~\cite{deitke2024molmopixmoopenweights}, keypoint detection tasks are formulated with the prompt \texttt{<KD>}, followed by a task-specific question. The output consists of class names and their corresponding points.

\vspace{1ex}\hspace{-3.5ex}\textbf{Open-Vocabulary Object Detection}\hspace{1mm} Object detection tasks are reformulated as open-vocabulary object detection. The prompt \texttt{<OVD>} precedes the target object class of interest, and the response includes the class name and its associated bounding boxes.

\input{ICCV2025-Author-Kit-Feb/figures/locations}

%% file: ICCV2025-Author-Kit-Feb/figures/locations.tex
\begin{figure*}[t!]
    \centering
    \includegraphics[width=\textwidth]{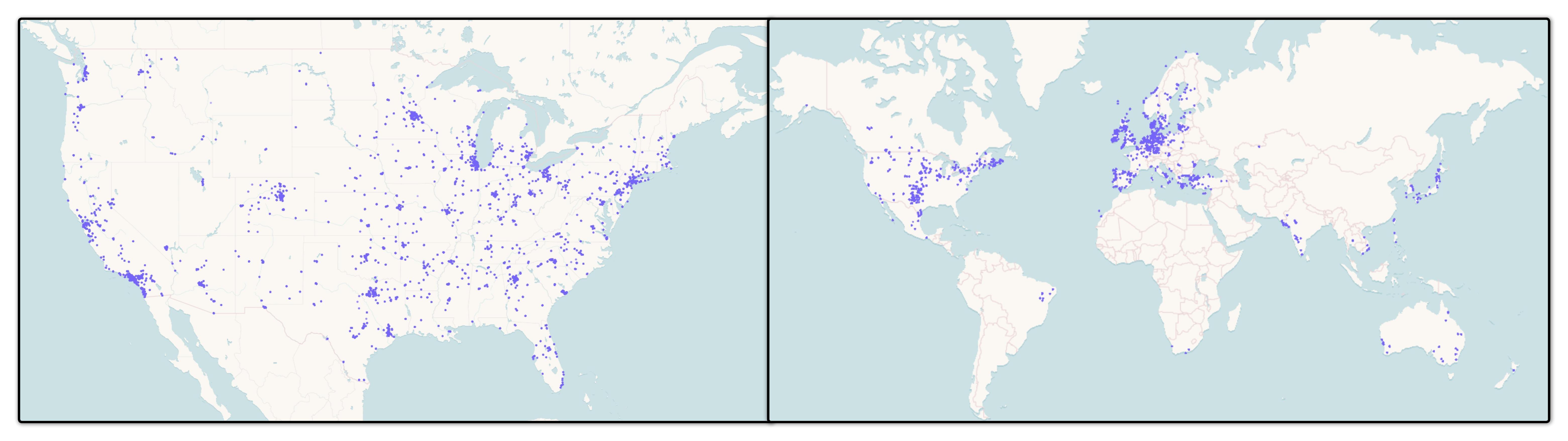}
    \caption{\textbf{Diverse geographic locations of the industrial assets in our InspectMM dataset.} InspectMM consists of imagery from inspections of properties \& solar assets (left) and wind (right). For properties and solar we sample 97k images from inspections across the continental U.S.; for wind we draw 145k images from from turbine inspections across the globe.}
    \label{fig:locations}
\end{figure*}

%% file: ICCV2025-Author-Kit-Feb/sec/data.tex
\section{DATASET}
\label{sec:data}

To train a unified model to perform industrial inspections across asset classes, image modalities, and visual tasks, we curate the InspectMM dataset. The dataset consists of 292,341 images taken during aerial drone inspections and 694,905 region-level annotations labeled by industrial inspection experts. The dataset, detailed in Table~\ref{tab:data}, consists of 3 visual tasks: image classification, object detection, and keypoint detection for 3 asset types: solar, wind, and properties.

\vspace{1ex}\hspace{-3.5ex}\textbf{Anomaly Flagging}\hspace{1mm} This subset of the dataset consists of imagery from wind turbine surfaces and building rooftops. This task requires global-level image understanding of whether an anomaly, which can vary from a large to small area, is present within a high-resolution image. The wind turbine imagery captures a variety of structural surfaces, including blades and hubs, while the buildings imagery spans both commercial and residential structures.

\vspace{1ex}\hspace{-3.5ex}\textbf{Anomaly Detection}\hspace{1mm} For anomaly detection, we use the ZVCD wind turbine crack detection dataset proposed in~\cite{agrawal2024barely}, and improve it by adding bounding box annotations, resulting in the ZVCD+ dataset. In this case, due to the small size of the cracks, the dataset is composed of small $1024 \times 1024$ patches of the original image. This adds complexity for a multitask model to be performant on both high-resolution imagery and patches.

\vspace{1ex}\hspace{-3.5ex}\textbf{Inventory Counting}\hspace{1mm} This subset consists of imagery extracted from a mixture of RGB and NIR orthomosaics of building rooftops and solar farms and is annotated with point-level coordinates for identifying and counting components such as HVAC units, vents, skylights, and solar panels. This method allows for efficient identification of objects of varying sizes and orientations without the complexity of detailed bounding box or polygon representations. Keypoint localization simplifies and lessens the annotation cost by focusing on identifying and marking the centroid of each component.

\input{ICCV2025-Author-Kit-Feb/tables/dataset}

%% file: ICCV2025-Author-Kit-Feb/tables/dataset.tex
\begin{table}[ht!]
\centering
\resizebox{1.0\linewidth}{!}{
\begin{tabular}{ccccc}
\toprule

\textbf{\begin{tabular}[c]{@{}c@{}}Inspection\\Task\end{tabular}} &
\textbf{\begin{tabular}[c]{@{}c@{}}Asset\\Type\end{tabular}} &
\textbf{\begin{tabular}[c]{@{}c@{}}Annotation\\Type\end{tabular}} &
\textbf{\# Images} &
\textbf{\# Annotations} \\

\toprule
\rowcolor{Gray}
\begin{tabular}[c]{@{}c@{}}Anomaly\\Flagging\end{tabular} & \begin{tabular}[c]{@{}c@{}}Properties\\Wind\end{tabular} & \texttt{CLS} & 145k & 145k \\

\begin{tabular}[c]{@{}c@{}}Anomaly\\Detection\end{tabular} & Wind & \texttt{OD} & 50k & 20k \\

\rowcolor{Gray}
\begin{tabular}[c]{@{}c@{}}Inventory\\Counting\end{tabular} & \begin{tabular}[c]{@{}c@{}}Properties\\Solar\end{tabular} & \texttt{KD} & 97k & 674k \\

\bottomrule
\end{tabular}%
}
\caption{\textbf{Overview of the InspectMM dataset for multitask multimodal drone inspections.} Industrial assets include wind turbines, solar farms, and residential and commercial buildings. Inspection tasks include anomaly flagging, anomaly detection, and inventory counting. The tasks can be mapped to the following machine learning problem types, respectively: Image Classification (\texttt{CLS}), Object Detection (\texttt{OD}), and Keypoint Detection (\texttt{KD}).}
\label{tab:data}
\end{table}

%% file: ICCV2025-Author-Kit-Feb/sec/experiments.tex
\section{EXPERIMENTS}
\label{sec:experiments}

\subsection{Zero-Shot}
We initially select the Florence-2~\cite{xiao2024florence}, GroundingDINO~\cite{liu2023grounding}, and PaliGemma~\cite{beyer2024paligemma} models as candidate VLMs. We use the ZVCD+ wind turbine crack detection dataset~\cite{agrawal2024barely} as our benchmark to evaluate the zero-shot performance of each VLM. As detailed in Table~\ref{tab:zero-shot}, we find that Florence-2 outperforms other VLMs with respect to the number of parameters and IoU. While GroundingDINO provides decent performance, its non-language-based decoder makes it complicated to adapt to multiple tasks other than open-vocabulary object detection. Therefore, we select Florence-2 as our architecture for the following experiments throughout.

\input{ICCV2025-Author-Kit-Feb/tables/results}

\subsection{Single \& Multitask Evaluation}
\textbf{Experimental Details}\hspace{1mm} We train the Florence-2 architecture and initialize the model weights from the original authors' checkpoints~\cite{xiao2024florence}. We use the AdamW optimizer~\cite{loshchilov2017fixing} with a learning rate of $\alpha=1e-6$, a cosine annealing schedule, mixed-precision training, a batch size of 8, and resize images to $768 \times 768$. We train each experiment for 10 epochs.

\input{ICCV2025-Author-Kit-Feb/tables/zero-shot}

\vspace{1ex}\hspace{-3.5ex}\textbf{Baselines}\hspace{1mm} For comparison to traditional single-task computer vision models, we use ResNet-50~\cite{he2016deep}, Faster R-CNN~\cite{ren2016faster} with a ResNet-50 Feature Pyramid Network (FPN) backbone, and Keypoint R-CNN~\cite{he2017mask} with a ResNet-50 Feature Pyramid Network (FPN) backbone for the Anomaly Flagging, Anomaly Detection, and Inventory Counting experiments, respectively, each initialized with ImageNet pretraining weights.

%% file: ICCV2025-Author-Kit-Feb/tables/results.tex
\begin{table*}[t!]
\centering
\resizebox{0.75\linewidth}{!}{
\begin{tabular}{cccccccc}
\toprule

\textbf{Task} &
\textbf{Type} &
\textbf{Model} &
\textbf{\# Params (M)} &
\textbf{Accuracy} &
\textbf{Precision} &
\textbf{Recall} \\

\toprule
\multirow{2}{*}{Anomaly Flagging} & \multirow{2}{*}{\texttt{CLS}} & ResNet-50~\cite{he2016deep} & 24 & 66.7 & 62.1 & 59.3 \\

& & InspectVLM (Ours) & 232 & \textbf{75.9} & \textbf{73.6} & \textbf{89.5} \\

\midrule

\multirow{2}{*}{Anomaly Detection} & \multirow{2}{*}{\texttt{OD}} & Faster R-CNN~\cite{ren2016faster} & 42 & - &\textbf{46.1} & \textbf{43.7} \\
& & InspectVLM (Ours) & 232 & - & 16.5 & 19.8 \\

\midrule

\multirow{2}{*}{Inventory Counting} & \multirow{2}{*}{\texttt{KD}} & Keypoint R-CNN~\cite{he2017mask} & 59 & - &\textbf{36.6} & \textbf{68.9} \\
& & InspectVLM (Ours) & 232 & - & 34.1 & 60.7 \\

\bottomrule
\end{tabular}%
}
\caption{\textbf{Results across the sub-tasks within our industrial asset inspection dataset, InspectMM.} For Anomaly Flagging classification we report overall accuracy, precision, and recall. For the Anomaly Detection and Inventory Counting we report precision and recall at a 50\% IoU threshold. Faster R-CNN and Keypoint R-CNN both use a ResNet-50 FPN backbone.  The tasks can be mapped to the following machine learning problem types: Image Classification (\texttt{CLS}), Object Detection (\texttt{OD}), and Keypoint Detection (\texttt{KD}). \textit{Note that InspectVLM is trained for all tasks simultaneously while each model comparison (e.g. ResNet-50) can only be trained on individual tasks.}}
\label{tab:results}
\end{table*}

%% file: ICCV2025-Author-Kit-Feb/tables/zero-shot.tex
\begin{table}[ht!]
\centering
\resizebox{0.8\linewidth}{!}{%
\begin{tabular}{lcccccc}
\toprule

\textbf{Method} & 
\textbf{\# Params (B)} &
\textbf{IoU} &
\textbf{mAP} \\

\toprule
PaliGemma~\cite{chen2023pali} & 3.00 & 0.00 & 0.00 \\
GroundingDino~\cite{ren2024grounding} & 0.31 & 0.47 & \textbf{4.48} \\
Florence-2~\cite{xiao2024florence} & \textbf{0.23} & \textbf{0.54} & 3.11 \\

\bottomrule
\end{tabular}%
}
\caption{\textbf{Zero-shot performance of multimodal language models on the ZVCD test set \cite{agrawal2024barely}}. We report box IoU and mAP as metrics to evaluate zero-shot performance in addition to model size in parameters.}
\label{tab:zero-shot}
\vspace{-3ex}
\end{table}

%% file: ICCV2025-Author-Kit-Feb/sec/results.tex
\section{DISCUSSION}
\label{sec:discussion}

\textbf{Domain Specific Datasets}\hspace{1mm} Due to VLMs being trained on natural images to be general-purpose models, zero-shot performance tends to be inadequate when transferring to domains like industrial inspection, which contain out-of-distribution imagery from the original pretraining set. As a result, their ability to generalize off-the-shelf to tasks such as anomaly detection, component identification, and inventory counting is limited. To address this gap, we find it necessary to construct a large-scale domain-specific dataset with high-quality annotations from human inspectors, like InspectMM, for fine-tuning VLMs through to performing accurate inspections across diverse asset types. Furthermore, we anecdotally find many openly available industrial inspection datasets are inadequate due to poor labeling quality or easily identifiable defects.

\vspace{1ex}\hspace{-3.5ex}\textbf{Experimental Results}\hspace{1mm} The results for the 3 tasks in our InspectMM dataset are presented in Table \ref{tab:results}. For Anomaly Flagging, our Florence-2 based InspectVLM significantly outperforms the traditional ResNet-50 classifier by 10\%+ precision and recall. For Inventory Counting, Florence-2 and Keypoint R-CNN achieve comparable performance. However, for object detection, we find that the VLM performs significantly worse than the Faster R-CNN model. We intuit that this is due to the nature of the ZVCD+ dataset being difficult with barely visible cracks requiring the extraction of fine-grained visual features which the VLMs haven't become capable of.

\vspace{1ex}\hspace{-3.5ex}\textbf{VLMs for Inspections}\hspace{1mm} Vision-language models (VLMs) offer a strong platform for multi-task learning, demonstrating versatility across a wide range of image domains and applications. Their ability to quickly adapt to new tasks, such as keypoint detection, makes them an appealing choice for dynamic environments where flexibility is crucial. However, traditional models like ResNet consistently achieve peak performance within their specific tasks, particularly when fine-tuned for specialized applications. This highlights a tradeoff between the benefits of streamlined model architecture and deployment, which VLMs provide, and the superior, task-specific performance that traditional models deliver. While VLMs excel at handling a variety of tasks, traditional models remain a viable choice for achieving high performance in narrowly defined problems. This tradeoff underscores the need for careful consideration when selecting models, depending on whether the goal is broader adaptability or maximum performance in a specialized context.

\subsection{Object Detection Failure Modes}

\input{ICCV2025-Author-Kit-Feb/figures/anomaly_detection}

While InspectVLM shows modest zero-shot capabilities and performs competitively on classification tasks, its object detection performance is significantly worse than traditional detectors. In this section, we analyze the model’s failure modes on the ZVCD+ subset of InspectMM, which requires detecting fine-grained cracks on wind turbine surfaces—arguably one of the most visually difficult and safety-critical inspection tasks in the dataset.

Rather than producing a spectrum of plausible detections, InspectVLM exhibits a trichotomy of failure modes: it either predicts inaccurate bounding boxes, overly large degenerate boxes, or hallucinates false defects that do not exist.

\subsubsection{Bounding Box Type Categorization}
We categorize each prediction from InspectVLM on the ZVCD+ test set into four groups:

\begin{itemize}
    \item \textbf{Accurate}: Predicted IoU $\geq0.5$ with a ground truth crack box.

    \item \textbf{Overly Predicted}: IoU $<0.2$ and the predicted box covers more than $30\%$ of the image area.

    \item \textbf{Inaccurate}: IoU $<0.5$ and the predicted box covers less than $30\%$ of the image area.

    \item \textbf{False Positive}: Falsely identified anomalies that do not correspond to any ground truth boxes.
\end{itemize}

Figure \ref{fig:categories_pie} shows the distribution of these results across the entire evaluation set. This distribution reflects a failure to achieve reliable localization - InspectVLM either guesses too broadly or fails to respond.

\subsubsection{Bounding Box Area Distributions}

To better understand this issue, we plot the relative size of predicted boxes normalized by the image size. Figure \ref{fig:area_dists} shows the normalized box area distributions for both InspectVLM predictions and ground truth annotations.

This mismatch confirms that the model often produces bounding boxes that are spatially incoherent, likely as a result of failing to learn proper attention mappings from prompt to image.

Figure \ref{fig:samples} shows representative examples of the three failure modes: an accurate detection with a tight bounding box, an overprediction where the model outputs a box spanning the majority of the image, and a missing prediction despite a clearly annotated crack. These examples illustrate that even when visual cues are present, the model either lacks the resolution to localize small cracks or defaults to a fallback decoding strategy that is only weakly grounded in the visual input.

\subsubsection{Limits of Visual Grounding in Current VLMs}

These object detection failure modes likely stem from several limitations inherent to current VLM architectures like Florence-2: Coarse spatial resolution from tokenized spatial embeddings may limit the model’s ability to attend to small visual structures. Decoder fallback behavior may default to predicting a large bounding box when confidence is low or grounding is ambiguous. Lack of multiscale visual processing, as used in traditional detectors (e.g., Feature Pyramid Networks), further weakens performance on small objects.

Moreover, because the ZVCD+ dataset contains real-world turbine surface conditions—variable lighting, textures, and scale—the model’s underperformance here highlights the gap between benchmark-style VLM pretraining and real industrial data.

\input{ICCV2025-Author-Kit-Feb/figures/samples}

\subsection{Structured vs. Unstructured Keypoint Detection}

The InspectMM dataset includes keypoint detection tasks for component counting across two domains: solar arrays and building rooftops. These tasks vary not only in object class (e.g., solar panels, HVAC units, vents), but also in spatial structure. Solar panels are typically arranged in regular, grid-like rows, while rooftop components are irregularly placed and visually diverse.

While the overall performance of InspectVLM trails traditional models, we find that it performs notably better on structured layouts. This suggests that VLMs may leverage spatial priors from pretraining or internal representations to guide keypoint localization in predictable scenes.

\subsubsection{Performance by Layout Type}

\input{ICCV2025-Author-Kit-Feb/tables/keypoint_performance}

To quantify this, we divide the keypoint validation set into: Structured layouts: Scenes containing solar panels in rows or grids. Unstructured layouts: Scenes with scattered rooftop fixtures.

We evaluate precision and recall at a 20-pixel distance threshold, comparing InspectVLM with a task-specific Keypoint R-CNN baseline. Results are shown in Table \ref{tab:kd-structured-unstructured}.

InspectVLM's performance in structured layouts is within 2–3\% of Keypoint R-CNN, but in unstructured layouts, it lags by over 20\% in both precision and recall.

Figure \ref{fig:samples} presents visualizations of InspectVLM predictions for both layout types: a solar array scene with well-aligned predicted keypoints, a rooftop with scattered HVAC units and partial or missed detections, and a rooftop with false positives placed in empty regions.

These examples demonstrate that the model is able to “fill in” keypoints along spatially regular patterns, even in cases of partial occlusion or shadowing. However, when objects do not follow predictable spatial arrangements, the model lacks sufficient visual sensitivity to local textures or edges to correctly localize them.

\subsection{Overfitting to Low Textual Variability}

Despite the promise of VLMs as flexible unified interfaces for vision tasks, we observe a critical failure mode in InspectVLM. The model easily overfits to fixed prompt structures and low-diversity answer spaces. This issue is particularly evident in the binary visual question answering (VQA) formulation of anomaly flagging, where prompts and answers are highly templated. Rather than grounding responses in image content, the model memorizes linguistic patterns and ignores visual evidence—resulting in performance collapse after a few epochs.

\subsubsection{Answer Frequency and Collapse}

\input{ICCV2025-Author-Kit-Feb/figures/answer_collapse}

We analyze the distribution of predicted answers across all tasks. Although the ground truth annotations contain a variety of (\texttt{yes}, \texttt{no}), bounding boxes, and keypoints, we observe that the model response distribution suddenly collapses. As shown in Figure \ref{fig:yes_percentage}, by epoch 4, all responses break down to \texttt{yes}, regardless of the ground truth or task grounding. This finding indicates that the model learns to optimize loss by memorizing the dominant label-response mapping, effectively ignoring the image modality.

\subsubsection{When Language Overpowers Vision}

These findings raise concerns about the core assumption behind treating vision tasks as language: while VLMs offer architectural unification, they can behave as language-only systems when the task setup encourages shortcut learning. In our anomaly flagging task, a combination of: Fixed prompts, Limited answer space, Class imbalance, and Repetitive training examples led the model to disregard its visual input and converge on a degenerate “always-yes” response.

This behavior highlights a serious limitation for applying VLMs in critical industrial contexts. When standardizing prompt formats for deployment (as one would in a production inspection pipeline), we may inadvertently create brittle models that appear accurate on validation data but fail to generalize to even slight variations in prompt phrasing or domain conditions.

We argue that prompt standardization, while desirable for consistency, should be accompanied by: prompt variation during training (e.g., paraphrased prompts), answer diversification (e.g., explanations or references), visual grounding checks (e.g., requiring spatial justifications), and language entropy regularization to penalize degenerate outputs.

\vspace{1ex}\hspace{-3.5ex}\textbf{Limitations \& Future Work}\hspace{1mm}
Due to the explosion in the number of VLMs being developed each week, it is impossible to compare to state-of-the-art architectures efficiently. Furthermore, we note that VLMs also excel at visual grounding tasks such as referring segmentation or object detection. However, in this paper, we do not evaluate VLMs for these industrial inspection tasks as creating these datasets is costly. We leave both of these for future work.

%% file: ICCV2025-Author-Kit-Feb/figures/anomaly_detection.tex
\begin{figure*}[!t]
    \centering
    \begin{subfigure}[b]{0.45\textwidth}
        \centering
        \includegraphics[width=\textwidth]{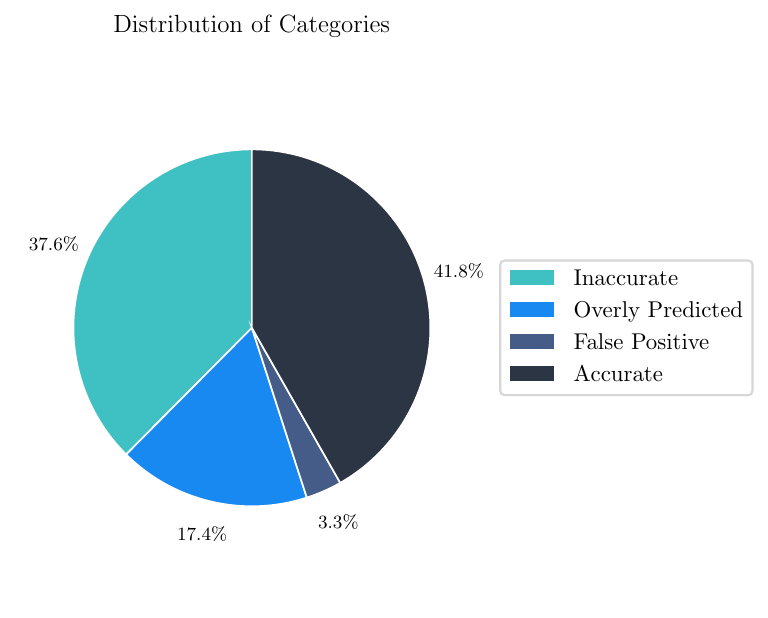}
        \caption{Distribution of detection quality categories, showing the proportion of accurate detections versus various error types.}
        \label{fig:categories_pie}
    \end{subfigure}
    \hspace{0.05\textwidth} %
    \begin{subfigure}[b]{0.45\textwidth}
        \centering
        \includegraphics[width=\textwidth]{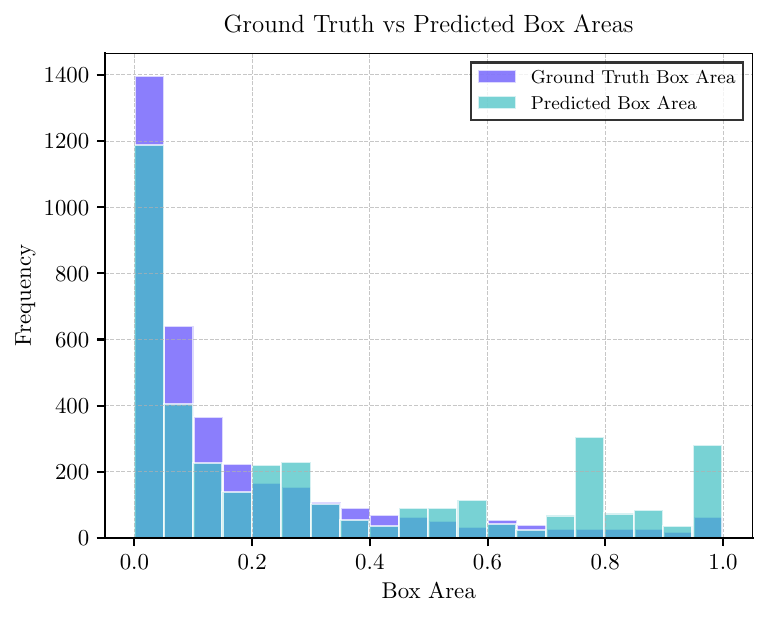}
        \caption{Distribution of ground truth vs. predicted normalized bounding box areas.}
        \label{fig:area_dists}
    \end{subfigure}
    \caption{Object detection performance metrics. The histogram in (a) displays the proportion of different detection quality categories, while (b) shows the distribution of ground truth and predicted bounding box areas normalized by image size.}
    \label{fig:combined_metrics}
\end{figure*}

%% file: ICCV2025-Author-Kit-Feb/figures/samples.tex
\begin{figure*}[t]
    \centering
    \includegraphics[width=0.95\textwidth, trim={0 5pt 0 60pt}, clip]{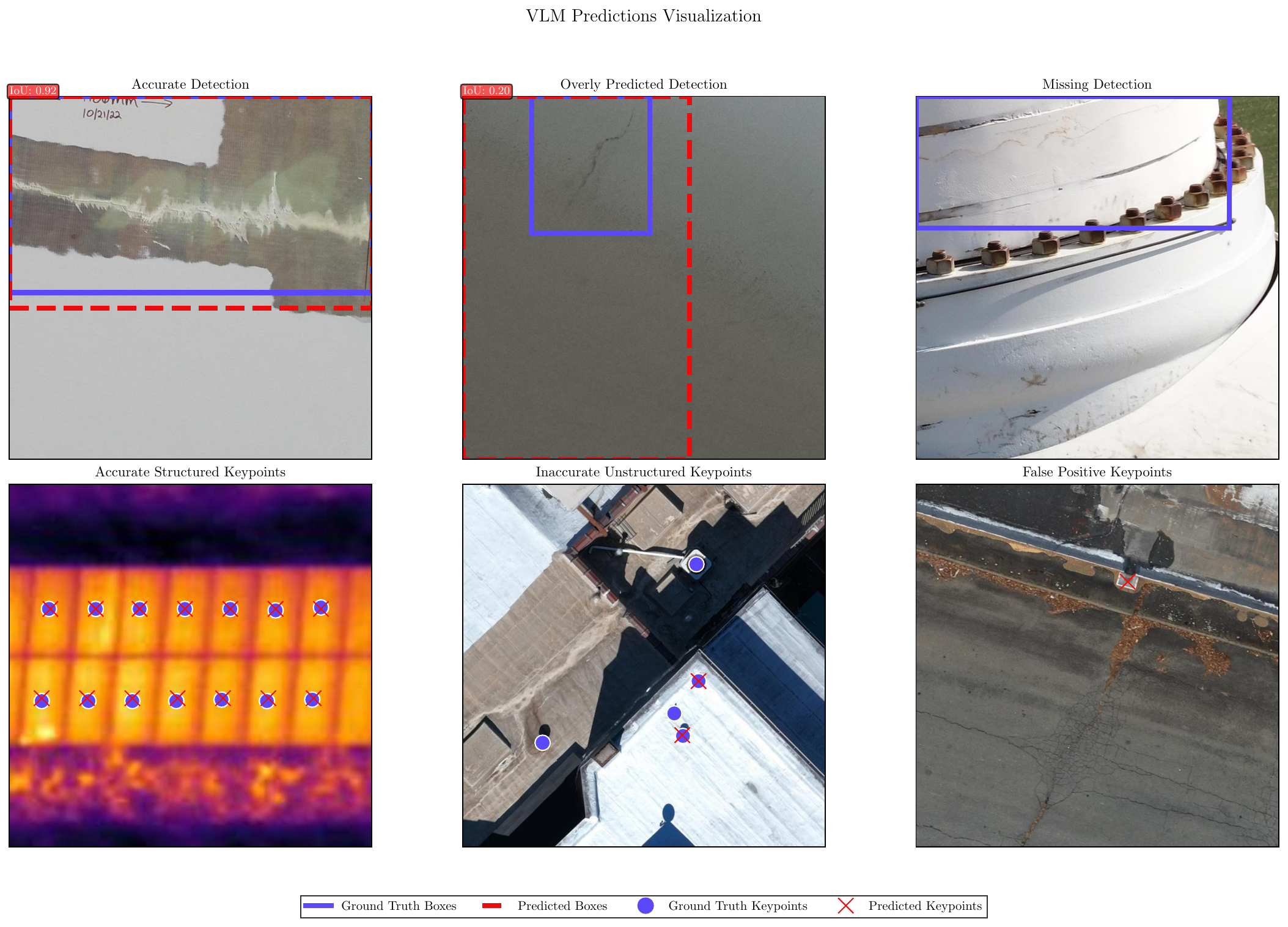}
    \caption{\textbf{Sample predictions from our InspectVLM on object and keypoint detection tasks.} Example behavior from the unified inspection model. Top: object detection behavior can be highly accurate (left), over-localized spanning the entire image (center), or misses important defects (right). Bottom: keypoint detection accuracy is largely based on structured data (left), or unstructured data (center and right).  \textit{Best visualized while zoomed in.}}
    \label{fig:samples}
\end{figure*}

%% file: ICCV2025-Author-Kit-Feb/tables/keypoint_performance.tex
\begin{table}[ht]
\centering
\resizebox{1.0\linewidth}{!}{
\begin{tabular}{lcccc}
\toprule
\textbf{Layout Type} & \textbf{Model} & \textbf{Precision (\%)} & \textbf{Recall (\%)} \\
\midrule
\multirow{2}{*}{Structured}   & InspectVLM      & 94.5 & 92.2 \\
                              & Keypoint R-CNN  & 97.7 & 91.6 \\
\midrule
\multirow{2}{*}{Unstructured} & InspectVLM      & 58.4 & 48.5 \\
                              & Keypoint R-CNN  & 69.1 & 73.9 \\
\bottomrule
\end{tabular}
}
\caption{Keypoint detection performance comparison for structured and unstructured layouts. We report precision and recall at a 10-pixel threshold. InspectVLM performs competitively in structured scenes like solar arrays but significantly underperforms in unstructured rooftop environments.}
\label{tab:kd-structured-unstructured}
\end{table}

%% file: ICCV2025-Author-Kit-Feb/figures/answer_collapse.tex
\begin{figure}[t!]
    \centering
    \includegraphics[width=0.98\linewidth]{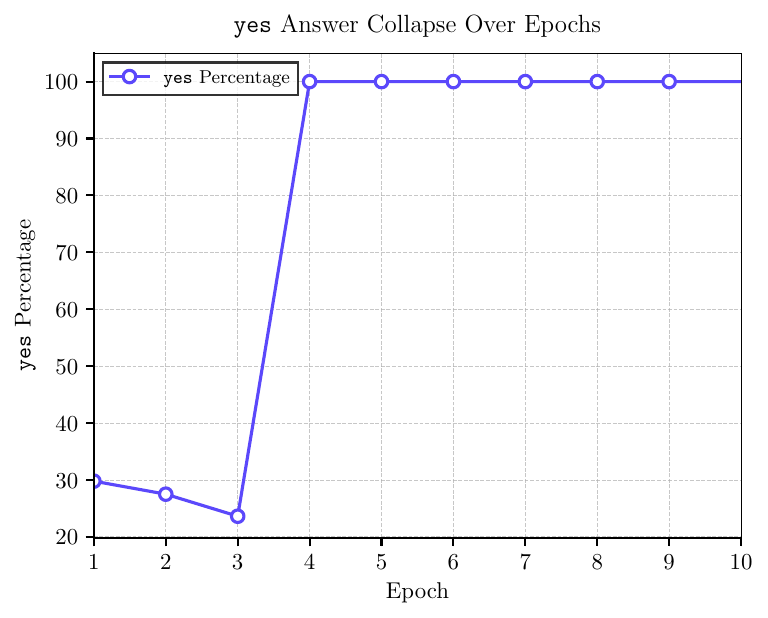}
    \caption{Evolution of \texttt{yes} responses across training epochs. The plot shows a clear pattern where the model initially maintains relatively low levels of \texttt{yes} responses (around 20-30\%) for the first few epochs, but then exhibits a collapse to 100\% \texttt{yes} responses starting from epoch 4 onwards.}
    \label{fig:yes_percentage}
\end{figure}

%% file: ICCV2025-Author-Kit-Feb/sec/conclusion.tex
\section{CONCLUSION}
\label{sec:conclusion}

In this work, we introduced InspectVLM, a vision-language model designed for multimodal, multitask industrial asset inspections. Leveraging the InspectMM dataset, we demonstrated that VLMs can unify traditionally independent inspection tasks, including anomaly detection, keypoint localization, and inventory management, with a single architecture. Our empirical results highlight the strengths of VLMs in multitask adaptability and language-based task unification, offering a streamlined alternative to maintaining separate task-specific models.

However, our findings also underscore key limitations. While VLMs exhibit competitive performance in anomaly flagging and keypoint detection, they struggle with fine-grained object detection, where specialized models like Faster R-CNN still outperform them. This suggests that while VLMs provide a flexible and scalable framework for industrial inspections, task-specific architectures remain essential for high-precision, domain-specific tasks.

%% file: main.bbl
\begin{thebibliography}{20}
\providecommand{\natexlab}[1]{#1}
\providecommand{\url}[1]{\texttt{#1}}
\expandafter\ifx\csname urlstyle\endcsname\relax
  \providecommand{\doi}[1]{doi: #1}\else
  \providecommand{\doi}{doi: \begingroup \urlstyle{rm}\Url}\fi

\bibitem[Agrawal et~al.(2024)Agrawal, Corley, Wallace, Vaughn, and Lwowski]{agrawal2024barely}
Sourav Agrawal, Isaac Corley, Conor Wallace, Clovis Vaughn, and Jonathan Lwowski.
\newblock Barely-visible surface crack detection for wind turbine sustainability.
\newblock In \emph{2024 IEEE/RSJ International Conference on Intelligent Robots and Systems (IROS)}, pages 5933--5939. IEEE, 2024.

\bibitem[Beyer et~al.(2024)Beyer, Steiner, Pinto, Kolesnikov, Wang, Salz, Neumann, Alabdulmohsin, Tschannen, Bugliarello, et~al.]{beyer2024paligemma}
Lucas Beyer, Andreas Steiner, Andr{\'e}~Susano Pinto, Alexander Kolesnikov, Xiao Wang, Daniel Salz, Maxim Neumann, Ibrahim Alabdulmohsin, Michael Tschannen, Emanuele Bugliarello, et~al.
\newblock Paligemma: A versatile 3b vlm for transfer.
\newblock \emph{arXiv preprint arXiv:2407.07726}, 2024.

\bibitem[Chen et~al.(2023)Chen, Wang, Beyer, Kolesnikov, Wu, Voigtlaender, Mustafa, Goodman, Alabdulmohsin, Padlewski, et~al.]{chen2023pali}
Xi Chen, Xiao Wang, Lucas Beyer, Alexander Kolesnikov, Jialin Wu, Paul Voigtlaender, Basil Mustafa, Sebastian Goodman, Ibrahim Alabdulmohsin, Piotr Padlewski, et~al.
\newblock Pali-3 vision language models: Smaller, faster, stronger.
\newblock \emph{arXiv preprint arXiv:2310.09199}, 2023.

\bibitem[Corley et~al.(2024)Corley, Lwowski, and Najafirad]{corley2024zrg}
Isaac Corley, Jonathan Lwowski, and Peyman Najafirad.
\newblock Zrg: A dataset for multimodal 3d residential rooftop understanding.
\newblock In \emph{Proceedings of the IEEE/CVF Winter Conference on Applications of Computer Vision}, pages 4635--4643, 2024.

\bibitem[Dai et~al.(2023)Dai, Li, Li, Tiong, Zhao, Wang, Li, Fung, and Hoi]{instructblip}
Wenliang Dai, Junnan Li, Dongxu Li, Anthony Tiong, Junqi Zhao, Weisheng Wang, Boyang Li, Pascale Fung, and Steven Hoi.
\newblock Instruct{BLIP}: Towards general-purpose vision-language models with instruction tuning.
\newblock In \emph{Thirty-seventh Conference on Neural Information Processing Systems}, 2023.

\bibitem[Deitke et~al.(2024)Deitke, Clark, Lee, Tripathi, Yang, Park, Salehi, Muennighoff, Lo, Soldaini, Lu, Anderson, Bransom, Ehsani, Ngo, Chen, Patel, Yatskar, Callison-Burch, Head, Hendrix, Bastani, VanderBilt, Lambert, Chou, Chheda, Sparks, Skjonsberg, Schmitz, Sarnat, Bischoff, Walsh, Newell, Wolters, Gupta, Zeng, Borchardt, Groeneveld, Nam, Lebrecht, Wittlif, Schoenick, Michel, Krishna, Weihs, Smith, Hajishirzi, Girshick, Farhadi, and Kembhavi]{deitke2024molmopixmoopenweights}
Matt Deitke, Christopher Clark, Sangho Lee, Rohun Tripathi, Yue Yang, Jae~Sung Park, Mohammadreza Salehi, Niklas Muennighoff, Kyle Lo, Luca Soldaini, Jiasen Lu, Taira Anderson, Erin Bransom, Kiana Ehsani, Huong Ngo, YenSung Chen, Ajay Patel, Mark Yatskar, Chris Callison-Burch, Andrew Head, Rose Hendrix, Favyen Bastani, Eli VanderBilt, Nathan Lambert, Yvonne Chou, Arnavi Chheda, Jenna Sparks, Sam Skjonsberg, Michael Schmitz, Aaron Sarnat, Byron Bischoff, Pete Walsh, Chris Newell, Piper Wolters, Tanmay Gupta, Kuo-Hao Zeng, Jon Borchardt, Dirk Groeneveld, Crystal Nam, Sophie Lebrecht, Caitlin Wittlif, Carissa Schoenick, Oscar Michel, Ranjay Krishna, Luca Weihs, Noah~A. Smith, Hannaneh Hajishirzi, Ross Girshick, Ali Farhadi, and Aniruddha Kembhavi.
\newblock Molmo and pixmo: Open weights and open data for state-of-the-art vision-language models, 2024.

\bibitem[Ding et~al.(2022)Ding, Xiao, Codella, Luo, Wang, and Yuan]{ding2022davit}
Mingyu Ding, Bin Xiao, Noel Codella, Ping Luo, Jingdong Wang, and Lu Yuan.
\newblock Davit: Dual attention vision transformers.
\newblock In \emph{European conference on computer vision}, pages 74--92. Springer, 2022.

\bibitem[Gu et~al.(2024)Gu, Zhu, Zhu, Chen, Tang, and Wang]{Gu_Zhu_Zhu_Chen_Tang_Wang_2024anomalyGPT}
Zhaopeng Gu, Bingke Zhu, Guibo Zhu, Yingying Chen, Ming Tang, and Jinqiao Wang.
\newblock Anomalygpt: Detecting industrial anomalies using large vision-language models.
\newblock \emph{Proceedings of the AAAI Conference on Artificial Intelligence}, 38\penalty0 (3):\penalty0 1932–1940, 2024.

\bibitem[He et~al.(2016)He, Zhang, Ren, and Sun]{he2016deep}
Kaiming He, Xiangyu Zhang, Shaoqing Ren, and Jian Sun.
\newblock Deep residual learning for image recognition.
\newblock In \emph{Proceedings of the IEEE conference on computer vision and pattern recognition}, pages 770--778, 2016.

\bibitem[He et~al.(2017)He, Gkioxari, Doll{\'a}r, and Girshick]{he2017mask}
Kaiming He, Georgia Gkioxari, Piotr Doll{\'a}r, and Ross Girshick.
\newblock Mask r-cnn.
\newblock In \emph{Proceedings of the IEEE international conference on computer vision}, pages 2961--2969, 2017.

\bibitem[Kumar et~al.(2024)Kumar, Alam, Farahat, Somineni, and Gupta]{kumar2024diagnostics}
Aman Kumar, Mahbubul Alam, Ahmed Farahat, Maheshjabu Somineni, and Chetan Gupta.
\newblock Diagnostics-llava: A visual language model for domain-specific diagnostics of equipment.
\newblock In \emph{Annual Conference of the PHM Society}, 2024.

\bibitem[Liu et~al.(2023{\natexlab{a}})Liu, Li, Wu, and Lee]{llava}
Haotian Liu, Chunyuan Li, Qingyang Wu, and Yong~Jae Lee.
\newblock Visual instruction tuning.
\newblock In \emph{Advances in Neural Information Processing Systems}, pages 34892--34916. Curran Associates, Inc., 2023{\natexlab{a}}.

\bibitem[Liu et~al.(2023{\natexlab{b}})Liu, Zeng, Ren, Li, Zhang, Yang, Li, Yang, Su, Zhu, et~al.]{liu2023grounding}
Shilong Liu, Zhaoyang Zeng, Tianhe Ren, Feng Li, Hao Zhang, Jie Yang, Chunyuan Li, Jianwei Yang, Hang Su, Jun Zhu, et~al.
\newblock Grounding dino: Marrying dino with grounded pre-training for open-set object detection.
\newblock \emph{arXiv preprint arXiv:2303.05499}, 2023{\natexlab{b}}.

\bibitem[Loshchilov et~al.(2017)Loshchilov, Hutter, et~al.]{loshchilov2017fixing}
Ilya Loshchilov, Frank Hutter, et~al.
\newblock Fixing weight decay regularization in adam.
\newblock \emph{arXiv preprint arXiv:1711.05101}, 5, 2017.

\bibitem[Ren et~al.(2016)Ren, He, Girshick, and Sun]{ren2016faster}
Shaoqing Ren, Kaiming He, Ross Girshick, and Jian Sun.
\newblock Faster r-cnn: Towards real-time object detection with region proposal networks.
\newblock \emph{IEEE transactions on pattern analysis and machine intelligence}, 39\penalty0 (6):\penalty0 1137--1149, 2016.

\bibitem[Ren et~al.(2024)Ren, Jiang, Liu, Zeng, Liu, Gao, Huang, Ma, Jiang, Chen, et~al.]{ren2024grounding}
Tianhe Ren, Qing Jiang, Shilong Liu, Zhaoyang Zeng, Wenlong Liu, Han Gao, Hongjie Huang, Zhengyu Ma, Xiaoke Jiang, Yihao Chen, et~al.
\newblock Grounding dino 1.5: Advance the" edge" of open-set object detection.
\newblock \emph{arXiv preprint arXiv:2405.10300}, 2024.

\bibitem[Wallace et~al.(2023)Wallace, Corley, and Lwowski]{wallace2023solar}
Conor Wallace, Isaac Corley, and Jonathan Lwowski.
\newblock Solar panel mapping via oriented object detection.
\newblock In \emph{ICLR 2023 Workshop on Tackling Climate Change with Machine Learning}, 2023.

\bibitem[Wang et~al.(2024)Wang, Li, Luo, Zhu, Yang, Rong, and Wang]{Wang_Li_Luo_Zhu_Yang_Rong_Wang_2024powerllava}
Jiahao Wang, Mingxuan Li, Haichen Luo, Jinguo Zhu, Aijun Yang, Mingzhe Rong, and Xiaohua Wang.
\newblock Power-llava: Large language and vision assistant for power transmission line inspection.
\newblock \emph{2024 IEEE International Conference on Image Processing (ICIP)}, page 963–969, 2024.

\bibitem[Xiao et~al.(2024)Xiao, Wu, Xu, Dai, Hu, Lu, Zeng, Liu, and Yuan]{xiao2024florence}
Bin Xiao, Haiping Wu, Weijian Xu, Xiyang Dai, Houdong Hu, Yumao Lu, Michael Zeng, Ce Liu, and Lu Yuan.
\newblock Florence-2: Advancing a unified representation for a variety of vision tasks.
\newblock In \emph{Proceedings of the IEEE/CVF Conference on Computer Vision and Pattern Recognition}, pages 4818--4829, 2024.

\bibitem[Zhu et~al.(2024)Zhu, Chen, Shen, Li, and Elhoseiny]{minigpt4}
Deyao Zhu, Jun Chen, Xiaoqian Shen, Xiang Li, and Mohamed Elhoseiny.
\newblock Mini{GPT}-4: Enhancing vision-language understanding with advanced large language models.
\newblock In \emph{The Twelfth International Conference on Learning Representations}, 2024.

\end{thebibliography}
